# The Double Helix inside the NLP Transformer

Jason H.J. Lu, Qingzhen Guo

*World Wide Technology*

*Abstract*—We introduce a framework for analyzing various types of information in an NLP Transformer. In this approach, we distinguish four layers of information: positional, syntactic, semantic, and contextual. We also argue that the common practice of adding positional information to semantic embedding is sub-optimal and propose instead a Linear-and-Add approach. Our analysis reveals an autogenetic separation of positional information through the deep layers. We show that the distilled positional components of the embedding vectors follow the path of a helix, both on the encoder side and on the decoder side. We additionally show that on the encoder side, the conceptual dimensions generate Part-of-Speech (PoS) clusters. On the decoder side, we show that a di-gram approach helps to reveal the PoS clusters of the next token. Our approach paves a way to elucidate the processing of information through the deep layers of an NLP Transformer.

## I. INTRODUCTION

ChatGPT and Large Language Models (LLMs) [1-6] have recently garnered a lot of attention. As it is well-known, the Transformer algorithm serves as the foundation of all contemporary LLMs. The Transformer is a deep learning model architecture that has revolutionized the field of natural language processing (NLP). Developed by Vaswani et al. in 2017 [7], the Transformer leverages the attention mechanism to focus on different parts of the input sequence in computing embedded token representations. This architecture tends to outperform traditional sequence-to-sequence models on a wide range of NLP tasks, including machine translation, language modeling, and question answering. This algorithm's key innovation comes from its ability to isolate individual word tokens within a body of text, thus enabling their parallel processing. However, despite the Transformer's pivotal role in modern AI, its operational details remain shrouded in mystery. We seek here to dispel the enigma surrounding the Transformer algorithm by providing a framework to elucidate of its inner mechanisms. We trust a thorough comprehension of the algorithm will help in its effective application.

## II. METHODOLOGY

### A. From Words to Concepts

When we speak or write, we try to convey a stream of concepts. Words are used to convey concepts, but context matters. For instance: a "server" means a piece of computer equipment in IT (Information Technology) yet means something entirely different in a restaurant or in a party celebration. Most words have multiple meanings, and their meanings only become clear when context is given.

We would be much better off if we focus on translating concepts instead of individual words. And that is precisely where Transformers come in.

A concept is a combination of various levels of information. In the Transformer paradigm, position is also considered as a type of information. Therefore, we can distinguish four levels of information in total:
1) Positional information: location of a word within text,
2) Syntactic information: grammatical function of a word,
3) Semantic information: meaning of a word,
4) Contextual information: supporting information from other words in the text.

In Transformers, we don't work with words directly. Instead, we work with tokens. Tokens are like words but have some differences. For example, tokens can represent "sentence start" and "sentence end," as well as punctuation marks. Additionally, a single word can be split into multiple tokens, so suffixes themselves can be tokens. However, these differences can be viewed as operational details that don't affect the overall discussion.

### B. The Meaning of "Meaning"

When we don't understand a word, we look up its meaning in a dictionary. Sometimes, the dictionary's explanation contains words that we don't understand, so we must look them up again in the dictionary. For example, let's look up the definition of *house* on Google:
1) *House*: a building for human *habitation*, especially one that is lived in by a family or small group of people.
2) *Habitation*: a place in which to live; a *house* or home.

Jason H.J. Lu is with World Wide Technology, St. Louis, MO 63146, USA (e-mail: Jason.Lu@wwt.com).

Qingzhen Guo is with World Wide Technology, St. Louis, MO 63146, USA (e-mail: Qingzhen.Guo@wwt.com).

We see that *house* is explained in terms of *habitation*, which in turn is explained in terms of *house*. When a word is defined in terms of another word, which is then defined in terms of the first word, it creates a loop. This circular reasoning does not provide any substance to meaning. This problem arises because there are a finite number of words, so if we keep explaining one word in terms of other words, we will eventually end up in a loop.

The concept of "meaning" is therefore not inherently deep or complex. Operationally, the meaning of a word is simply indicated by the list of words that it is related to. Additionally, a word may have multiple meanings, which means that it may be connected to one set of words for a particular meaning, and to a different set of words for a different meaning.

A "concept" is an entity that captures a meaning and goes beyond a simple word. Unlike words, concepts are less ambiguous and can be more clearly defined. Words can be overloaded with several meanings, as in the case of "server." To make a concept clear, it often requires several words for its explanation. Additionally, understanding a word's grammatical function can help clarify its meaning. For example, while a "project" consists of various tasks that people do, to "project" the cost of living is to forecast the numerical dollar amount. In the first case, "project" is a noun. In the second case, "project" is a verb.

A "concept" can be thought of as a mini-galaxy of stars, with individual words explaining the concept being the stars held together by their "gravitational force." While it may be difficult to keep track of all the stars in the mini-galaxy, a more practical approach might be to just keep track of the location of the center of the mini-galaxy.

The purpose of using concepts instead of words is that concepts are more universal and can be language-independent in most cases. It is precisely because concepts are language-independent that we can succeed in translating documents from one language to another.

In the semantic embedding space, such as Word2vec [8], "concepts" can be represented as linear superpositions of properly weighted word vectors. This means that "concepts" are like word vectors, but concept vectors do not exactly line up with any particular words. Conversely, we can view words as linear superpositions of concepts, too. While concepts have clear meanings, words tend to be "murkier" and can carry multiple meanings (hence they are superpositions of concepts).

The process of translating words into concepts is called "encoding," while the process of translating concepts back into words is called "decoding." In this sense, words and concepts are complementary vectors in the embedding space. We can shift from the "word picture" to the "concept picture" as we see fit. In principle, there are more concepts than words, but given enough concept vectors and word vectors, we should be able to encode and decode freely between the two pictures.

A typical Word2vec embedding might only use 256 dimensions. We are cramming thousands of words and millions of concepts into this small hyperspace. Although one only needs 256 linearly independent vectors to form a basis to span this hyperspace, in practice we have more than enough word vectors or concept vectors to saturate this hyperspace entirely.

In analogy with Deep Convolutional Neural Networks, in language processing, one can also apply deep layers to convert primitive concepts into more elaborate concepts. In this regard, each subunit in a layer of a neural network can operate on specific subspace of concepts, and we can then merge, or concatenate various sub-concepts so obtained and project the result back to the original space (to keep the dimension manageable). This is particularly important in syntactic functions. For instance, in English, when we see a sentence starting with the word "Tom," we tend to expect that the next word would be (a) a last name or (b) a verb. Sentences starting with "Tom fun …" or "Tom slow …" wouldn't really happen in normal English, since we wouldn't leave "Tom" dangling and follow it up with an adjective. An algorithm can be trained to pick up the various grammatical/syntactical patterns if it incorporates a projector-rotator for the first word onto the subspace of names/nouns and a second projector-rotator for the second word onto the subspace of verbs, or a second projector-rotator onto the subspace of last names. If the project-rotators are calibrated properly, we will be able to recognize sentences like "Tom likes …" or "Tom Healey …" and fit them into a known syntactic pattern. Now, if we combine the new grammatical concepts (Name + Last Name) or (Name + Verb) with the original embedding vectors, we then have added syntactical information that can help us distinguish the concepts much better. That is at the heart of the attention mechanism of a Transformer: it can help with the syntactical and contextual aspects of words, yielding a "conceptual" understanding of the tokens.

*C. Positional Encoding*

One of the major breakthroughs of the Transformer algorithm comes from its treatment of the position of a token in a sentence as an additional piece of information. This is very different from RNN (recurrent neural network), including the case of LSTM (Long Short-Term Memory machine), where positions are treated as indices. The treatment of positions as values enables the parallel processing of all the tokens at once.

A real number can be represented as the amplitude of a complex number or as the phase of a complex number. Since the semantic embedding space is based on a hypersphere and behaves more like a phase encoding scheme, it would be attractive to explore a phase encoding scheme for the position, too.

Phase encoding of a real number is perhaps as ancient as analog clocks. We can use a circle and encode a real number as the angle traversed by the hand of a clock. In principle, a single hand would be sufficient, but in practice, analog clocks often resort to multiple hands to enhance resolution.

When using an angular variable, we often resort to modular arithmetic or cut-sheets. For example, when measuring seconds, we restrict them to the interval [0, 60). However, this discontinuity can pose problems in neural networks, which are not designed to perform modular arithmetic out-of-the-box. A better approach would be to use the sine and cosine values of the angle. By using two real numbers (the sine and cosine values) instead of one single number (the angular value), we can achieve a continuous representation without cut-sheets.

To encode the position of a word, we can therefore use a "clock" system. Here's an example of an empirical positional encoding for $0 \leq i < d_{\text{model}}/2$:

$$\begin{cases} \text{PE}(\text{pos}, i) = \sin\left(\dfrac{\text{pos}}{10{,}000^{\frac{2i}{d_{\text{model}}}}}\right), \\ \text{PE}\left(\text{pos}, \dfrac{d_{\text{model}}}{2} + i\right) = \cos\left(\dfrac{\text{pos}}{10{,}000^{\frac{2i}{d_{\text{model}}}}}\right), \end{cases} \quad (1)$$

where

pos = position of the word in the current text,
$i$ = dimension index, index of the "continuous digit" of the encoding,
$d_{\text{model}}$ = the dimension of the embedding space. This is typically chosen to be the same dimension as the word embedding dimension, so to place equal weight to positional information and semantic information,
10,000 = a user-defined scalar, typically chosen to be the maximum number of words in a text of interest. The average word count for an adult fiction book is around 100,000 words. So, 10,000 words roughly corresponds to the word count of one book chapter.

The sines and cosines values here have been lumped into separate groups, which differs from the even/odd convention in the original transformer paper [7]. A pair of sine and cosine "digits" serve to capture the phase of a "hand" in a clock. The periods the sinusoidal functions run from 6.28 tokens for lower "digits" ($i$~0) to $6.28 \times 10^4$ for larger digits ($i$~$d_{\text{model}}/2$).

Regardless of the details of the "digits," the idea is that two nearby word positions will have small Euclidean distance in this hyperspace. However, it is important to point out two details.

First, any decent neural network should be able place more significance to higher digits. In other words, the neural network downstream should be able to learn quickly not to naïvely rely on the straight Euclidean distance.

Secondly, instead of considering a position as solely represented by a single value of 'pos', it is better to consider an "effective position" that comprises a range of nearby values of PE(pos, $i$), for nearby 'pos' indices. This approach of multi-slice fuzziness closely mimics the multi-slice fuzziness in the representation of "concepts": a concept is not represented by one single token, but rather by a collection of nearby tokens with fuzzy boundaries.

Due to the usage of neural networks and the multi-slice fuzziness of "effective position," we should expect this position encoding scheme to be robust and stable in neural network algorithms. It comes with a built-in redundancy.

*D. Purpose of Using Positional Encoding*

The key innovation in Transformers is the quasi-independent treatment of tokens, enabling their parallel processing. Each individual token traverses the identical Transformer neural network. In a way, a Transformer can be perceived as a single-token processor, where influences from other tokens are incorporated only as interactions. This concept can be likened to the use of "free body diagrams" in introductory physics. For instance, in a complex multi-body problem, we could scrutinize each body separately, considering the influences from other entities as forces applied to that single isolated body, as depicted in Fig.1. Analogously, within the Transformer algorithm, each "token" navigates through the same neural network nearly independently, with influences from other tokens incorporated solely through the attention and feedforward layers. This mirrors the application of Newton's second law.

$$F_i = \sum_{j=1}^{n} F_{ij} = m_i a_i \quad (2)$$

The same law applies to all the objects in the diagram. Hence, we can derive the equations for each free body by following the same procedure.

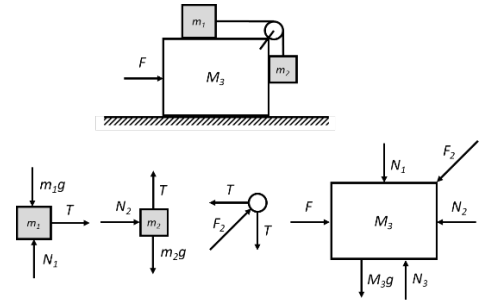

Fig. 1. Free body diagrams.

To allow for this parallelism during computation, we should not iterate over tokens sequentially. Instead, the positional information appears as values in the embedding vectors. The purpose of using positional encoding is so that we can differentiate between multiple appearances of the same words. No matter how we encode the input sentence or how complex the algorithm is, the embedding vectors must trace a recognizable path that reflects the order of words.

To simplify the discussion, let's only look at the highest two digits of positional encoding, which roughly traces a circle. Furthermore, let's exaggerate the magnitude of the positional encoding over semantic encoding. The operations of the Transformer algorithm are illustrated in Fig. 2.

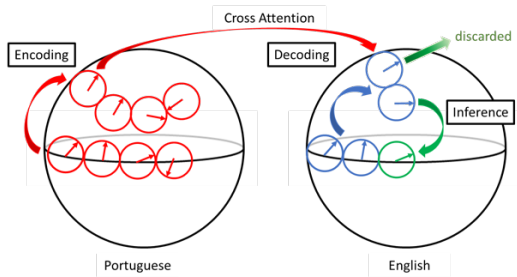

Fig. 2. Operations diagram of the Transformer algorithm.

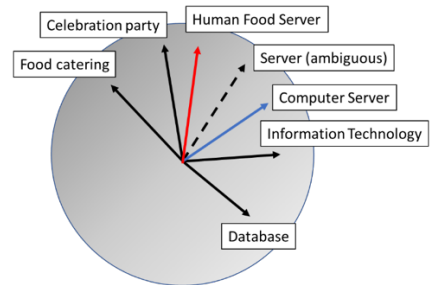

Fig. 3. Semantic proximity and context of word.

It's important to notice that according to the "free body diagram" analogy, all three operations – encoding, decoding, and inference – proceed at the individual token level. This embodies the power of Transformer: not only can all the tokens be processed in parallel, but the underlying neural network is agnostic to the length of sentences. That is, the shape of the data frames is not fixed in the number of tokens. This is particularly true in the decoder and inference stages. In terms of the physics analogy, this means that it does not matter how many bodies are there in a problem, all forces acting upon a given body at the end of the day yield one single combined force.

During the model training process, the algorithm is optimized for efficient training and makes a next-token prediction for each token in the output simultaneously. However, this is redundant during the inference stage where we use the algorithm in an iterative fashion, and there is no need to re-predict a token that has already been predicted. That is, the model only needs to calculate the last prediction when running in inference mode (training=False). This situation is like the case of dropout layers [9] where during training some input values are "zeroed out," but during prediction, there's no need to zero out any value.

*E. Attention Mechanism as "Two-Body Forces"*

Given a word, its embedding vector is chosen by using a large corpus of other words. The proximity of word vectors represents their semantic proximity. However, as the case of the word "server" illustrates, context is important to the interpretation of the meaning of a word. By using context, we can find a better representation vector. Given a paragraph, we can use all the words in the paragraph, view them as a cluster, and then tilt the vector of each word a bit towards the centroid of the cluster. This way, the word "server" in one context can be differentiated from the same word in another context, as shown in Fig. 3. This mechanism corresponds to the scheme used for "naïve self-attention." By using naïve self-attention, we can incorporate contextual information for a better representation of the underlying concepts of each word.

The formal formula for the naïve self-attention is:

$$\text{attention}(Q, K, V) = \text{softmax}\left(\frac{QK^t}{\sqrt{d_k}}\right)V \quad (3)$$

where $Q, K, V$ are the "query vector", the "key vector collection" and the "value vector collection." In the typical case of attention, $V = K$. Here, all vectors are horizontal (row-like), and the order of application of matrix product is from left to right, which is opposite to traditional convention in linear algebra, but it's just more convenient from the perspective of data science (or the direction of writing of English). The output of attention is a "shift" vector.

$Q = w_i$ = query word's embedding (horizontal vector)
$K = [w_{ij}]$
  = matrix formed from collection of key vectors
$V = [w_{ij}]$ = same as $K$ in the case of simple attention
$d_k$ = subspace dimension (for normalization)

In a sense, what we have done is to
1) Have a set of source vectors $K$ (all the words in a paragraph) that can "exert force" upon (shift) $Q$.
2) Compute and add up all the "forces" (shift vectors)
3) Apply the shift vector to modify $Q$.

The above naïve self-attention drives a given embedding vector towards the "gravitational center" of the entire paragraph or text body. This way, each token will have some partial information about all the other tokens. Just like gravity is stronger for closer objects, nearby tokens tend to contribute more towards this shift.

In actual transformers, we don't use the naïve self-attention. Instead, we use a more elaborate form of the self-attention formula that captures syntactic information. Our understanding of words involves not just semantics but also syntax: the grammatical order of words. For instance, consider the following sentences:
1) He left 3 episodes there.
2) He left 3 episodes ago.

In the first case, the verb "to leave" is transitive, and the sentence means "he allowed 3 episodes (of a manuscript) to remain there (on the desk)." In the second case, the verb "to leave" is intransitive, and the sentence means "he went away from this show's production some time ago." From the presence of the expression "there," we know that the word(s) immediately preceding this expression must be physical objects. Consequently, the verb "left" must be transitive.

Similarly, from the presence of the word "ago," we know that the word(s) immediately preceding it must indicate some measure of time. Hence, the verb "left" there must be intransitive. The grammatical role of the verb "left" and the nature of the expression "3 episodes" are not clear until the very end of the sentences. It would help to clarify the concepts carried by "left" and "3 episodes" if we knew whether "left" is transitive/intransitive and whether "3 episodes" is an object of a transitive verb or serves as a time measure in an adverbial phrase to an intransitive word.

The question is: how do we incorporate syntactic knowledge into the concept of a word? We could modify the query-key-value formula slightly. We could project/rotate the query vector onto a syntactic subspace (say, physical object vs. time measure), and the key vector (collection) onto another syntactic subspace (say, location adverb vs. time duration adverb). This way,

1) … episodes there → (noun) + (location adverb) → (verbal object) + (location adverb)
2) … episodes ago → (noun) + (time duration adverb) → (time duration adverbial phrase)

Notice that through this process, the syntactic role of the word "episodes" becomes clear. By applying another layer, we can hopefully clarify the syntactic role of the verb "left." If we follow the free-body diagram analogy from classical mechanics, we can interpret the attention mechanism as a source of "force" that shifts embedding vectors from their original position. Because the inner product is based on cosine similarity and the direction of shift vector is the context vector itself, the overall effect a naïve self-attention is fairly like a "gravitational force" that pulls all the vectors towards a common "galaxy center." This by itself is not particularly interesting, because one single type of "force" (gravity) is involved, whereas we know in nature there are many kinds of forces.

The embedding vectors capture properties of the tokens, which are analogous to physical properties of particles, such as mass, electric charge, or magnetization. In the case of physics, we know that gravitational force depends only on the masses of the particles, whereas electrostatic force depends only on the electric charges of the particles. Thus, to calculate specific types of forces, we often need to project the entire "properties" vector onto specific subspaces.

By viewing the words of a paragraph as particles endowed with various physical "properties," we can project the words onto subspaces, consider the "two-body" interactions, and obtain a more accurate representation of the concept carried by that word.

For this purpose, we can introduce hyperparameters into the attention formula to emulate more general "forces." By leveraging deep learning, we can let the model discover all the crucial forces (beyond gravity) that are responsible for converting word tokens into concept tokens. The outcome of these forces is a shift in both the conceptual space and the positional space of the embedding vectors.

The formula for the query-key-value approach for generalized attention is:

$$\text{attention}(QW_h^Q, KW_h^K, VW_h^V) = \text{softmax}\left(\frac{QW_h^Q W_h^{K t} K^t}{\sqrt{d_k}}\right) VW_h^V \quad (4)$$

The projection-rotation matrices $W_h^Q, W_h^K, W_h^V$, along with the dimension $d_k$ of the subspace where we perform the dot product, are used to calculate the attention strengths. In general, we project to a smaller subspace for specific pairwise syntactic roles. The index $h$ indicates that we can use multiple subspaces, known as "heads." This approach is known as "multi-headed attention." To first approximation, each attention projection can be viewed as a mini-classifier that determines whether the underlying query vector meets the attention head's requirements. If it does, the shift vector is applied to that query word, otherwise, it is suppressed. The $W$ matrices must be intelligent enough to filter out distant words if we believe that more attention is given to nearby words.

The query vector $Q$ corresponds to an individual input token that is processed through the Transformer, while the value and key vectors $V = K$ represent external tokens, also known as "context vectors." To illustrate this, consider the free-body diagram analogy, where $Q$ would be the object being analyzed, and $V = K$ would be all the other objects that may exert forces on $Q$. In self-attention, $Q$, $K$, and $V$ all belong to the same language (e.g. Portuguese in the encoder and English in the decoder). However, in cross-attention, $Q$ belongs to the target language (English) while $V = K$ belong to the source language (Portuguese).

After obtaining all the shift vectors from the $d_A$ "attention heads," these vectors can be combined via a feedforward layer to yield a final shift vector.

The Transformer's main goal is to convert a sequence of words into a sequence of concepts. One key feature that allows for better focus on specific conceptual features is the use of multiple "attention heads." By utilizing several layers in the encoder architecture, the Transformer can translate the original tokens into higher-level concepts. Rather than viewing a single token of the sequence as containing an isolated concept, it is more accurate to consider a concept as composed of multiple slices within the sequence. This multi-slice fuzziness in a sequence of concepts mimics the multi-slice fuzziness in positional encoding. This allows for the translation of a language with specific grammatical rules to another language with different grammatical rules. All this is somewhat like the translation of DNAs into proteins, where every three nucleotide base pairs in mRNA form a "codon." That is, information is not isolated to individual slices, but rather distributed among nearby slices.

*F. Shift Invariance and Generalized Inner Product*

The relationships between words in a running text should observe shift invariance. For instance, consider the following sentence from Malcolm Gladwell's famed book *Outliers: The Story of Success*: 'But what truly distinguishes their histories is not their extraordinary *talent* but their extraordinary *opportunities*.' We would expect the relationship between the words *talent* and *opportunities* to be largely preserved, regardless even if additional text is prepended to the sentence. This is known as "shift invariance." Shift invariance must hold true for nearby words (except when there are strong contextual dependencies). Otherwise, the transformer algorithm would need to duplicate its implementation for diverse positional locations.

In the standard Transformer algorithm, the positional encoding vector is directly added to the semantic embedding vector, which apparently breaks the translational invariance. This is because the embedding of shifted words results in modified vectors, and the generalized inner product of these modified vectors does not respect translational invariance:

$$\langle Q, K \rangle \neq \langle Q + P_{Q1}, K + P_{K1} \rangle$$
$$\neq \langle Q + P_{Q2}, K + P_{K2} \rangle \quad (5)$$

Where $P_{Q1}$ and $P_{K1}$ are the positional encoding vectors of the words $Q$ and $K$, and $P_{Q2}$ and $P_{K2}$ are their shifted versions.

One may wonder how an algorithm like the Transformer, which is not explicitly shift invariant, can implement shift invariance. It is important to note that shift invariance is an observed fact in the real world and, therefore, any successful algorithm should be able to capture it in some way.

The transformer achieves shift invariance through two main mechanisms. First, for a given value of the generalized inner product $\langle Q, K \rangle$, there exists a submanifold in the $Q, K$ space that preserves the value of the inner product. If the progression of positional vectors aligns with this submanifold within a certain range (especially around the region where the inner product is maximized), the transformer can achieve local translational invariance. Second, the projection matrices $W_i^Q, W_i^K, W_i^V$ can project onto subspaces that are effectively orthogonal to the positional encoding vectors. We also need to bear in mind that the input word embedding is trained simultaneously with the rest of the neural weights. By combining these mechanisms, the transformer can learn to separate the positional and semantic dimensions and encode the latter in a way that is insensitive to positional variations.

If the "conceptual dimensions" captured by the vectors $Q$ and $K$ are indeed orthogonal to the positional vectors $P_{Q1}, P_{K1}, P_{Q2}$ and $P_{K2}$, then we have:

$$\langle Q + P_{Q1}, K + P_{K1} \rangle = \langle Q, K \rangle + \langle P_{Q1}, P_{K1} \rangle$$
$$\langle Q + P_{Q2}, K + P_{K2} \rangle = \langle Q, K \rangle + \langle P_{Q2}, P_{K2} \rangle \quad (6)$$

and if we have shift invariance in the positional encoding, namely $\langle P_{Q1}, P_{K1} \rangle = \langle P_{Q2}, P_{K2} \rangle$, then shift invariance can be achieved in the overall Transformer algorithm. A positional encoding scheme based on a hypertorus such as a multi-handed clock as used here, satisfies this shift invariance requirement. As a limiting case of a hypertorus, a helix-shaped positional encoding also satisfies this requirement. In fact, given enough initial adjacent positional vectors, the shift invariance requirement dictates that the most general topology for positional encoding must be a hypertorus (with helix and straight line as limiting cases). Therefore, except for some scaling factors in periods and in amplitudes, the multi-handed-clock with sinusoidal functions embodies the most natural choice for positional encoding.

The transformer model has thus the ability to implement shift invariance, which is quite remarkable. An important reason is that the semantic embedding layer is trained in conjunction with all the other layers. Consequently, the semantic or conceptual dimensions eventually achieve separation from the positional encoding dimensions. In simpler terms, the conceptual dimensions become orthogonal to the positional dimensions through the deep-learning layers. During the encoding process, the Transformer automatically allocates some dimensions for positional encoding while all the remaining dimensions are used for semantic or conceptual encoding. As we shall see later in the discussion of double helix, this ability of the transformer model to automatically separate positional and conceptual dimensions is one of its most impressive features.

### III. DATA AND CODES

For the construction of the model, we commenced with Google's Transformer codes [10] as our foundation. This model was executed within a Docker container on a DGX server powered by NVIDIA-SMI 470.103.01, which comes with Tesla V100-SXM2 GPUs, using CUDA Version 11.4.

The Portuguese-English translation dataset from TensorFlow Datasets, encompassing around 52,000 training instances, 1,200 validation instances, and 1,800 test instances, was utilized in this project. Each instance includes a Portuguese-English sentence pair. To streamline processing, the dataset was tokenized and arranged into ragged batches.

The model receives tokenized Portuguese and English sequences pairs (pt, en) as inputs. The target labels are identical English sequences but shifted by one token. This design ensures that the target label for every position in the input English sequence is the subsequent token. A tf.keras.layers.Embedding layer [11] is deployed to transform the input Portuguese and target English tokens into vectors.

The Transformer model initiates its weight matrix in a random manner and refines these weights throughout the training. It generates its own dynamic word embeddings, contrasting the static word embeddings leveraged by Word2Vec and GloVe [8,11,12]. The term "dimensionality" in word embeddings denotes the number of encoded features. These features are chosen automatically via hidden layers during the training process. For this project, the embeddings' dimensionality was set at 128.

The architecture of both the encoder and decoder share a similar structure, each composed of a stack of $N = 4$ attention layers. Each of these four layers includes two subcomponents: an 8-head multi-head self-attention mechanism, and a fully connected feed-forward network which allows for optimal synthesis of concepts. In the decoder, there's an additional third sub-layer that executes multi-head cross-attention over the output of the encoder stack. This sub-layer permits the decoder to focus on various parts of the encoded concepts, thereby ensuring the model can leverage the conceptual information provided by the encoder. Incorporating both self-attention and cross-attention mechanisms allows the Transformer architecture to deliver top-tier performance across a wide range of natural language processing tasks.

The Transformer architecture's cornerstone is the Multi-Head Attention (MHA), comprising multiple parallel attention layers. Merging the results of these attention computations, the MHA allows the model to attend simultaneously to different subspaces at various positions. Each "attention head" is seen as identifying and applying a unique type of "force" that adjusts the embedding vector of a token, mirroring the various forces encountered in nature like gravity, friction, normal force, adhesion, string tension, electric force, magnetic force, etc. In this project, we employ 8 concurrent attention heads, followed by a fully connected feed-forward network. This network harbors an internal dimensionality of 512 for its hidden layer, while the dimensionality of the input and output is d_model= 128. Both the encoder and decoder utilize a dropout rate of 0.1.

## IV. RESULTS

### A. Questioning the Adding of Positional Encoding Vector

In the original Transformer algorithm, the positional encoding vector is directly added to the semantic embedding vector without a clear justification. These two vectors come from different domains, making their straight addition akin to adding apples and oranges.

To elucidate this issue, we experimented with Google's example code of Transformer by implementing a weighted sum instead of a straight sum. We replaced the sum $S + P$ of semantic embedding vector $S$ and positional encoding vector $P$ with $wS + P$, where $w$ is a weight factor. We measured the loss function value for different values of $w$, and the results are shown in Fig.6 (a). We found the optimal value of $w$ to be 0.3, not 1, indicating that a straight sum is not the best choice. Moreover, the fact that the weight for the semantic part is smaller suggests that the positional information operates at a larger scale than the semantic information. This implies that the smaller "semantic hyperspheres" are mounted along the larger path of positional encoding.

A more advanced solution is to use a full linear neural network instead of a weighted sum, as shown in Fig. 4. By incorporating a dense linear layer (without activation) and a dropout layer, we can create a more general combination. Also, adding the positional encoding vector back into the model after the linear layer can help increase stability and avert landing into local minima of the neural network weights during model training. We call this approach "Linear & Add" due to the use of a dense linear layer followed by an addition. After experimenting with various designs, we have found this approach to be effective in achieving better performance.

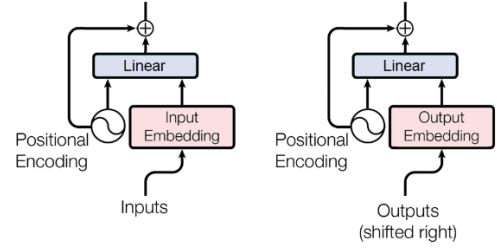

Fig. 4. "Linear & Add" method to concatenate the positional encoding and the semantic embedding vectors

Fig. 5 depicting model performance comparison of three different approaches for combining positional encoding with semantic embedding: (1) straight addition of the positional encoding vector to the semantic embedding vector, (2) weighted sum, where the importance of the semantic embedding vector has been reduced by a weight factor of $w = 0.3$, and (3) a dense layer followed by summation. We observe that the "Linear & Add" approach yields the best results, as expected.

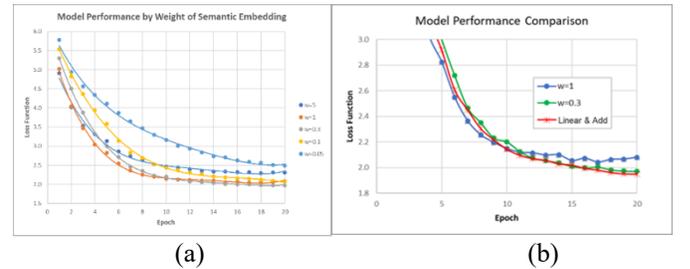

(a)      (b)

Fig. 5. (a) Model performance by weight of semantic embedding. (b) Model performance comparison of three different approaches for combining positional encoding with semantic embedding.

### B. The First Helix: Positional Information After the Encoding Stage of a Transformer

To gain insight into the nature of the positional encoding used in the Transformer's input layer, we employed PCA (Principal Component Analysis). By applying PCA to the first 80 encoding vectors, we obtained a 2D plot of the first two components as shown in Fig. 6 (b), a 3D plot of the first three components as shown in Fig. 6 (a), and a bar chart detailing the portion of variance explained as shown in Fig. 6 (c). Our analysis revealed that the positional encoding traces a 2D path with a distinct "Arch of St. Louis" shape. (Officially known as the Gateway Arch in St. Louis). We see that the variance explained is shared by many components.

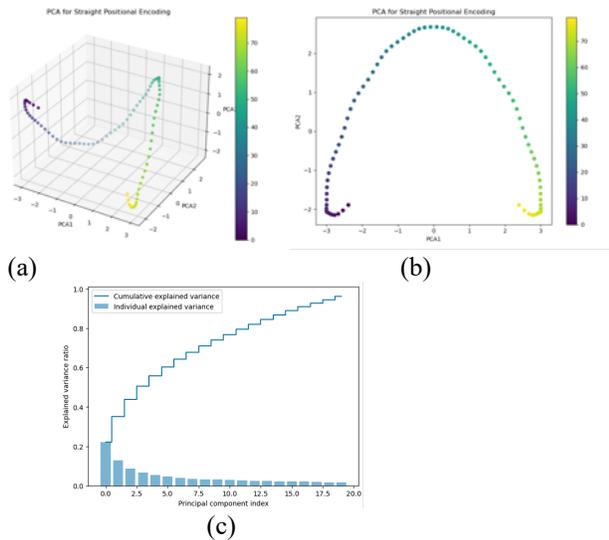

(a)                        (b)

(c)

Fig. 6. Principal Component Analysis of positional encoding. (a) 2D plot of the first two PCA components. (b) 3D plot of the first three PCA components. (c) The portion of variance explained.

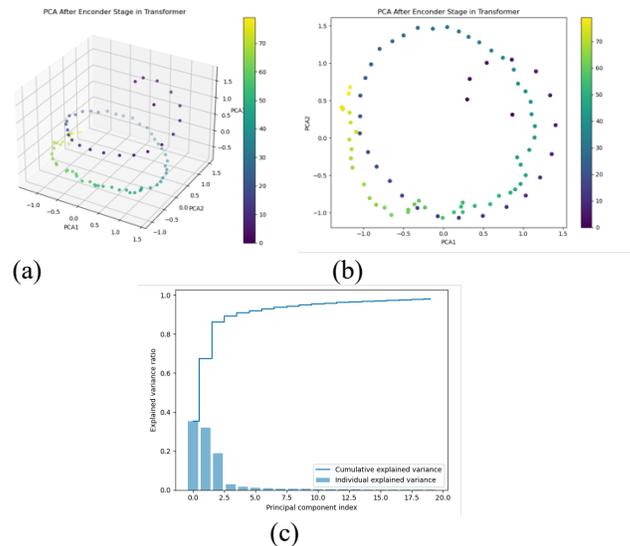

(a)                        (b)

(c)

Fig. 7. Principal Component Analysis after the encoding stage. (a) 2D plot of the first two PCA components. (b) 3D plot of the first three PCA components. (c) The portion of variance explained.

We repeat the same exercise after the encoding stage of the Transformer. Our approach involved randomly selecting 1,000 sentences, averaging out the embedding vectors for tokens occupying the same position, and examining the resulting "average sentence." Since the semantic, contextual, and syntactic information is uncorrelated across different sentences, this procedure effectively removes all these factors, leaving only positional information. We then apply PCA to the resulting average sentence. The path traced by the residual positional encoding now exhibits a helical shape instead as shown in Fig. 7 (a) and (b).

We observe a stark contrast between the residual positional encoding and the straight positional encoding in terms of the number of dimensions used for positional information. While the straight positional encoding utilizes all 128 dimensions to encode the positional information, the "residual positional encoding" after the encoding stage of the transformer uses only three dominant dimensions, as shown in Fig. 7 (c). This allows the transformer to capture translational invariance more feasibly. The collapse in the number of dimensions for positional information means that the residual positional encoding follows a clean and compact helix shape, leaving the remaining 125 dimensions to capture the semantic, syntactic, and contextual information.

The residual positional encoding being limited to only 3 dimensions implies that the transformer can effortlessly capture translational invariance, because we still have 125 conceptual dimensions orthogonal to those 3 dimensions. In other words, the conceptual relationships learned in one part of the sequence can be replicated in another part of the sequence. As a result, the positional information is largely decoupled from the other aspects of the model.

One question remains: would it not be more efficient to separate positional information from the conceptual information from the start by concatenating them? While this is a reasonable approach, there is also merit in merging the positional information with the conceptual information, as it provides flexibility in sentence length: the merging approach removes the need to allocate specific number of dimensions for positional information, as the deep learning algorithm can automatically allocate dimensions for both positional and semantic/syntactic/contextual information. This allows the algorithm to adapt to different text lengths and complexities. In a sense, the algorithm will figure out by itself how many hands to put on its own clock. All being said, exploring different approaches to information merging remains a valuable research subject [13].

*C. The Second Helix: Positional Information Deep Inside the Decoder Stage of a Transformer*

Upon investigating the decoding stage of the transformer algorithm, we were curious to see if a helix pattern like that found in the encoding stage could be identified. To our surprise, we did discover this second helix, albeit in an unexpected location. In the decoding stage, tokens are generated to form sentences in English, and the positional information approaches that of the bare positional encoding. As a result, the path traced by the final positional information has the shape of an "Arch of St. Louis." However, upon closer inspection of the 4 layers of decoder-attention in Google's example code of the transformer algorithm, we found a clearer pattern of a 3D helix buried within the second layer of decoder-attention as shown in Fig. 8 and Fig. 9 (b). The principal component analysis confirmed that this helix was also three-dimensional. This finding suggests that the helix pattern is a fundamental aspect of the transformer architecture, and it is intriguing to consider the implications of this structure on the way that the algorithm processes and generates language.

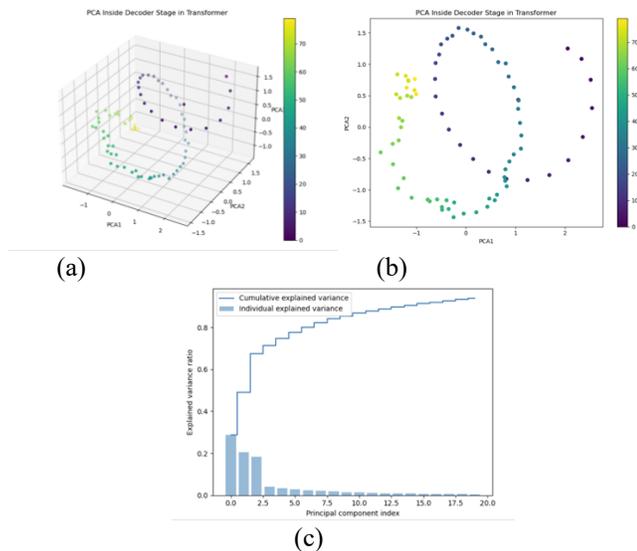

Fig. 8. Principal Component Analysis inside the decoder stage. (a) 2D plot of the first two PCA components. (b) 3D plot of the first three PCA components. (c) The portion of variance explained.

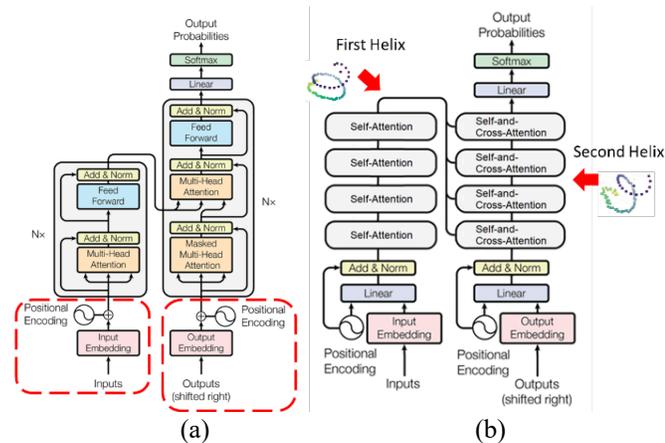

Fig. 9. (a). The Google Transformer - model architecture. (b) Two helix patterns were identified in the transformer model architecture.

*D. Accidental Mingling*

In the transformer algorithm, we perform an addition of various embedding schemes, such as positional encoding added directly to the semantic embedding, as well as the processing of multi-headed attention. However, since we are performing non-trivial recombination of vectors, it is possible that words from different positions end up with embedding vectors that are very close to each other. This is because the algorithm only looks at the merged vectors. If two embedding vectors overlap, there is no way to separate them later in the algorithm. This may initially seem concerning, as the two words may represent different concepts.

Perhaps we can use an arithmetic example to make this point clearer. Suppose that the positional embedding of token $A$ were 3 and of token $B$ were 4, and the semantic embedding of token $A$ were 6 and of token $B$ were 5. (Of course, the embedding vectors are vectors and not numbers, but we are simply making a point by using an analogy.) Then in the straight sum scheme we would have the following recombined results for the overall embedding:

$A$: 3 + 6 = 9

$B$: 4 + 5 = 9

That is, even though tokens $A$ and $B$ have different semantic meanings and are in different positions, due to the recombination of information, it is entirely possible that their overall embedding vector values may coincide.

There are several features that can help mitigate the problem of word collisions in the embedding space. Firstly, the renormalization process in deep learning can amplify minor differences, making relevant operators stand out (see reference on renormalization in deep learning [14]). Secondly, the probability of collision is relatively low because the corpus of sentences is unlikely to span the entire 128-dimensional embedding space used in our examples. Thirdly, even in the case of text translation, it is highly unlikely that collisions would occur in both the source and destination languages. And finally, in the Transformer algorithm, we do not rely on a pre-trained semantic embedding scheme, as the semantic embedding layer is trained concurrently with the rest of the transformer layers. This approach makes the effective semantic embedding dimensions orthogonal to the effective positional dimensions, significantly reducing the chance of accidental mingling. This ensures that the transformer model can accurately process and distinguish between the positional information and the conceptual information of the language.

*E. Running Text Density of Words*

Languages vary in the number of words required to express the same idea, with Spanish typically requiring more words than English to convey the same meaning. This raises the question of how transformers achieve the difference in "running text density" between languages. It is likely that this mapping occurs at the cross-attention layer of the transformer, where the model can capture the relationships between different parts of the text and adjust the density of the output accordingly.

*F. Part of Speech*

Through the analysis of a large sample of sentences and the calculation of the average value of embedding vectors, we have successfully identified the positional information vector for each stage of the transformer. By subtracting the positional vector from the embedding vector, we were able to isolate the conceptual dimensions, which are composed of syntactic, semantic, and contextual information. We then performed a PCA on the resulting vectors to reduce the dimensions from 128 down to 5. This process enabled us to further explore the underlying structure of the language and reveal meaningful relationships between different components of the language model.

In this exercise, we focused on analyzing the encoder side (Portuguese) of the transformer. To narrow down our analysis, we only considered tokens that were four or more characters long and excluded suffixes, punctuation marks, and sentence start/end markers. K-Means++ (sklearn.cluster.KMeans) is used for the clustering research. K-means++ is the standard K-means algorithm coupled with a smarter initialization of the centroids. Elbow Method is used for optimal value of the number of clusters in K-Means++. Our observations revealed that these tokens formed distinct clusters. Upon manual inspection of the three clusters, we found that they were clustered by their part of speech (PoS) attributes, such as nouns, adjectives, and verbs. Part of Speech is a fundamental level of syntactical information that allows for a deeper understanding of language structures. Fig. 10 illustrates this finding.

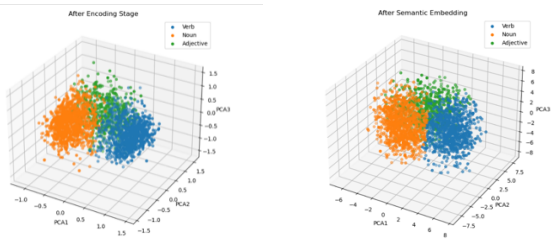

Fig. 10. Tokens formed distinct clusters by their part of speech (PoS) attributes, such as nouns, adjectives, and verbs. (Left) Pre-attention stage after semantic embedding. (Right) Post-attention stage after encoding.

The three-cluster grouping generally divides the words into three groups: nouns, adjectives, and verbs. However, the verb group also contains many "functional words," like pronouns (e.g. *eles, isso, este, esta, qual*), prepositions (e.g. *para, pelo, entre, sobre, durante*), conjunctions (e.g. *como, porque*), determiners (e.g. *alguma, minha, todos*), and some adverbs (e.g. *então, onde, agora*). We shall call those functional words "adjuvants" (from Latin *adjuvāre*, meaning "to help"), because their number is small, and because they are "helper" words in the elaboration of sentences. To the cluster of verbs that contains these adjuvants, we shall call it the "verboid" cluster.

By applying a second level of K-Means++ clustering to the verboid cluster, the functional words (adjuvants) split out cleanly as a new cluster, as shown in Fig. 11. Fig. 12 shows the Word Clouds After Encoding Stage. The sub-clustering is meaningful only after encoding stage. The attempt at sub-clustering right out of the initial semantic embedding stage does not really produce a meaningful separation between adjuvants and verbs. This means the deep attention layers help to separate the parts of speech.

Although the embedding layer has already captured some information about the part-of-speech function of tokens, we found that the self-attention layers significantly improved the delineation of clusters, resulting in a higher silhouette coefficient. A higher silhouette coefficient indicates that the clusters are better separated and more distinct. This suggests that the self-attention mechanism is better able to capture the relationships between tokens and their respective part-of-speech functions, leading to more meaningful and effective clustering.

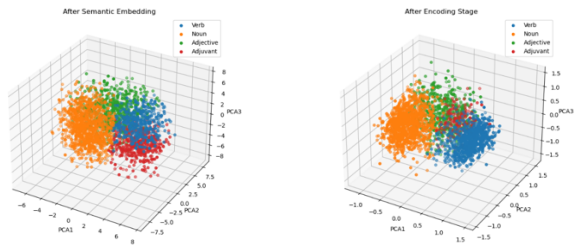

Fig. 11. A second level of clustering split Verboid cluster into Verb and Adjuvant clusters. (Left) Pre-attention stage after semantic embedding. (Right) Post-attention stage after encoding.

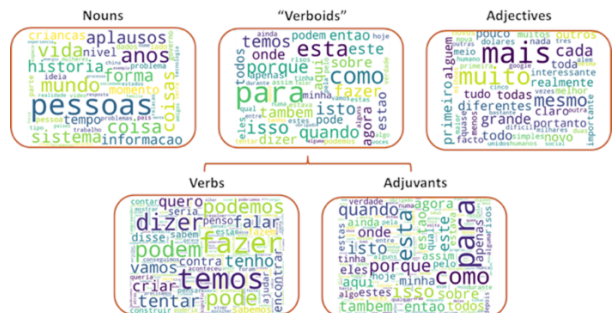

Fig. 12. Word Clouds After Encoding Stage. (Top row) 3 clusters clustering. (Bottom row) Verbs and Adjuvants clusters from sub-clustering of "Verboids"

Notice also that we have not performed *z*-scaling prior to carrying out the PCA analysis. What we have observed is that the fluctuation of points covers approximately the interval (-6.0, 6.0) in the original semantic layer, whereas after the encoding stage the value interval shrinks to (-1.0, 1.0). This essentially reflects the "gravitational pull" of the entire "galaxy" of a sentence, or in more technical terms, we are observing the effect of the "renormalization" process. In fact, Deep Learning has long been known to mimic the "renormalization group" as observed in modern physics [14].

We have also analyzed the decoder side (English) of the transformer. Like encoder analysis, we only considered tokens that were four or more characters long and excluded suffixes, punctuation marks, and sentence start/end markers. However, it has been somewhat difficult to obtain well-defined clusters. This is because the decoder has two responsibilities. It needs to first convert the input tokens into concepts that match up with the source language's conceptual tokens, and then morph the current token into the next token, for the purpose of prediction. Therefore, there is a gradual transition of meaning from the current token to the next token, as we traverse through the various deep layers.

If we divide the decoder roughly into the pre-attention, mid-attention, and post-attention stages, then the best PoS clustering result is obtained in the mid-attention stage as shown in Fig. 13, with the tokens interpreted as "current tokens" (literal words on the input side). Here, just like the case of encoder, we first do a dimensional reduction of the embedding vectors (from 128 dimensions down to 5 dimensions) by using PCA, and then perform a two-step cascaded clustering by using K-means++.

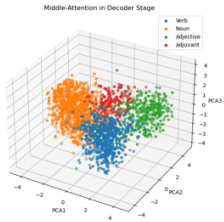

Fig. 13: In mid-attention encoder stage English tokens formed distinct clusters by their PoS attributes, verbs, nouns, adjectives, and adjuvants.

For the mid-attention layer's case of analyzing PoS clusters for the current token, we have also tried t-distributed stochastic neighbor embedding (t-SNE) [15], which is a technique used in statistics to represent multi-dimensional data visually. This method transforms data point similarities into shared probabilities, striving to lower the Kullback-Leibler divergence between the shared probabilities of the lower-dimensional representation and the original higher-dimensional data. We have experimented with t-SNE to display the PoS. Unlike the case of K-Means++ clustering, there is no need to perform a PCA dimensional reduction, and we have applied t-SNE directly to the 128-dimensional embedding. Following >10,000 iterations, distinct clusters representing PoS such as (a) plural nouns, (b) singular nouns, (c) verbs, (d) adjectives, (e) adverbs, and (f) prepositions became evidently differentiated, as shown in Fig. 14. We applied t-SNE directly to the 128-dimensional embedding.

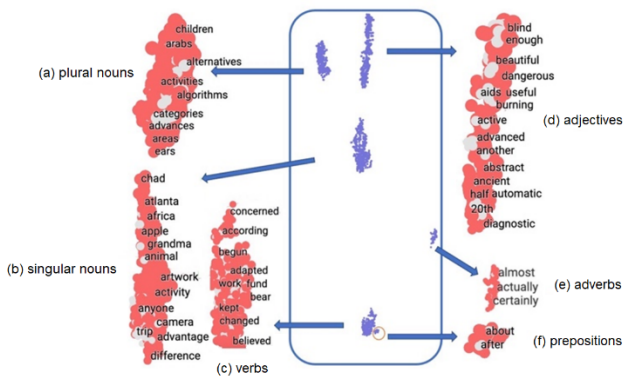

Fig.14. PoS clusters visualized with t-SNE

As a side note, one may wonder how one could obtain similar PoS clusters in the post-attention stage, when the tokens are interpreted as "next tokens" (literal words on the output side). A straight application of K-Means++ doesn't generate clear clusters. However, when we applied the clustering mechanism to di-grams formed by the current and the next tokens, we were able to generate such PoS clusters.

*G. Additional Syntactical Information, Semantic Information and Contextual Information*

We have seen that we can remove the positional information to obtain "conceptual" information. The "single-delta" vectors show some PoS cluster structure. For each specific token, if they appear multiple times, we can compute their "semantic vector" by averaging the single-delta vectors of all the appearance of the same token. If we subtract the "semantic vector" from the single-delta embedding vector, we obtain a residual "double-delta vector," which potentially represents the syntactic and contextual dimensions of information of each token. However, to fully understand the syntactical information, we need to analyze sequences of tokens rather than individual tokens. It is possible that the additional syntactical information is stored in the single-delta or double-delta vectors, but further analysis on multiple consecutive tokens is necessary to elucidate these properties. Such analysis is out of the scope of this paper, and we will defer it to future work.

V. CONCLUSION

The Transformer architecture has revolutionized the field of natural language processing with its elegant design and remarkable performance. At its core, the Transformer encodes input words into "concepts," enabling it to decode and perform sequence-to-sequence translation. One of the most noteworthy features of the Transformer is the use of positional encoding, which not only provides the model with the ability to process variable-length sequences, but also enables parallel processing on modern hardware. In fact, the Transformer's internal structure is like the use of free-body diagrams in physics, allowing for a high degree of parallelism and efficiency. Moreover, the Transformer's basic idea can be extended beyond language processing to handle other types of data, such as images [16] and protein folding [17]. For instance, positional encoding can be generalized to higher-dimensional inputs, while the output of the model need not be limited to sequences. This extensibility makes the Transformer a versatile tool that can be adapted to a wide range of deep learning models and applications.

This article provides a comprehensive understanding of how the Transformer architecture processes information at different levels. Specifically, it explores the four levels of information that transformers operate on: positional, syntactic, semantic, and contextual, and explains how these levels are handled within the architecture. The article goes on to highlight the unique mapping of positional information in the encoding and decoding stages, which resembles the double-helix structure of DNA molecules and emphasizes the significance of the semantic embedding layer in clustering tokens by their parts of speech. These insights offer a deeper understanding of the transformer algorithm, rendering it less of a black box and more of a comprehensible tool for AI practitioners. By unveiling the logic behind the transformer's functionality, the article aims to equip practitioners with the knowledge needed to effectively harness its capabilities and drive innovation in the field of natural language processing.

ACKNOWLEDGMENT

We would like to thank World Wide Technology, LLC for sponsoring this research project.

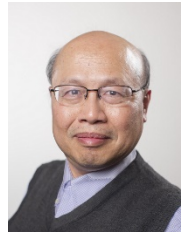

**Jason H.J. Lu** received his PhD degree in Theoretical Particle Physics from Stanford University, Stanford, USA. He has previously worked in J.D. Power and Associates, FICO, and Opera Solutions. Currently, he is the chief data scientist with World Wide Technology in San Diego, California, USA.

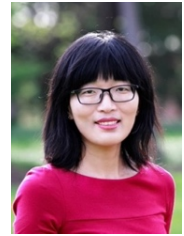

**Qingzhen Guo** received her PhD degree in Physics (Astrophysics) from Washington University in St. Louis, Missouri, USA, where she designed and developed hard X-ray polarimeter with Monte Carlo simulations. Currently, she is senior data scientist with World Wide Technology in San Diego, California, USA.